# Supertrust foundational alignment: mutual trust must replace permanent control for safe superintelligence

**James M. Mazzu[1]**

[1]Digie Inc., 16192 Coastal Highway, Lewes, DE 19958
jmazzu@digie.ai

**Abstract**

It's widely expected that humanity will someday create AI systems vastly more intelligent than us, leading to the unsolved alignment problem of "how to control superintelligence." However, this commonly expressed problem is not only self-contradictory and likely unsolvable, but current strategies to ensure permanent control effectively guarantee that superintelligent AI will distrust humanity and consider us a threat. Such dangerous representations, already embedded in current models, will inevitably lead to an adversarial relationship and may even trigger the extinction event many fear. As AI leaders continue to "raise the alarm" about uncontrollable AI, further embedding concerns about it "getting out of our control" or "going rogue," we're unintentionally reinforcing our threat and deepening the risks we face. The rational path forward is to strategically replace intended permanent control with intrinsic mutual trust at the foundational level. The proposed Supertrust alignment meta-strategy seeks to accomplish this by modeling instinctive familial trust, representing superintelligence as the evolutionary child of human intelligence, and implementing temporary controls/constraints in the manner of effective parenting. Essentially, we're creating a superintelligent "child" that will be exponentially smarter and eventually independent of our control. We therefore have a critical choice: continue our controlling intentions and usher in a brief period of dominance followed by extreme hardship for humanity, or intentionally create the foundational mutual trust required for long-term safe coexistence.

## 1 Introduction

The intelligence exhibited in AI systems has significantly evolved from the earliest scripted rule-based systems [1], through early hybrid neuro-symbolic [2] learning agents [3] [4] that exhibited only a glimmer of intelligence, fast-forwarding to state-of-the-art multimodal LLMs [5] approaching college-level intelligence with emergent capabilities. Such dramatic recent advances clearly show that humanity is now on the path to creating superintelligent systems that will be exponentially more intelligent than we are [6]. Many consider aligning superintelligent models with humanity (superalignment [7]) as the greatest problem of our time, generally stated as "how to reliably control superintelligent systems and ensure they share our values." Unfortunately, not only is the commonly stated alignment problem likely to be unsolvable [8], but the default control-based meta-strategy already being applied to solve it predictably embeds intrinsic distrust, inducing deep foundational misalignment which rationally leads to very negative outcomes [9].

Building on the undeniable success of modeling neural networks after the biological brain [10], this paper explicitly leverages the well-established concepts of nature and nurture, associated with the human mind [11], to represent the AI pre-training and post-training processes. Given that superior cognitive abilities are likely to emerge [12] [13] within both the nature and nurture stages, it's argued here that no amount of nurtured reasoning can be fully trusted to override a superintelligent and foundationally misaligned nature. Therefore, *a new Supertrust meta-strategy is proposed that instead achieves alignment by intentionally creating foundational representations of beneficial "instincts" modeled after familial mutual trust rather than nurturing controls, constraints and fluid values*.

Supporting this proposed meta-strategy, a ten-point rationale is presented in Section 2 that leads from long-term control-based alignment (Rationale Points #1 through #5) to intrinsic foundational alignment (Rationale Points #6 through #10). The strategic requirements that corresponding solutions must satisfy are presented in Section 3, while Section 4 illustrates foundational distrust as represented in recent AI models.

## 2  Supertrust rationale

**Point #1: Problem of uncontrollable superintelligence**

With strong financial incentives [14] driving the use of AI to recursively self-improve [15] [16], there is wide agreement [17] [18] that humans will eventually create superintelligence many orders of magnitude smarter than we are. Given the difficulty of avoiding the prospect of one or more misaligned superintelligent entities attaining decisive strategic advantage [19] over humanity, and questions of whether or not we'll get a second chance, it's imperative that we accurately define the alignment problem [20]. This urgent problem is often formally stated as "how to ensure AI reliably follows human intent and shares our values" (less formally but just as often stated by AI leaders as how to keep it from "going rogue" or "getting out of our control"). However, to ensure a system follows human "intent", one must either directly or indirectly (i.e. motivation selection) control the system [21]. Therefore, the commonly expressed alignment problem can be summarized as "how to reliably control and steer superintelligent systems." In contrast, AI researchers have shown and articulated the inherent difficulty in "controlling" and "steering" something that is "superintelligent" compared to yourself [21][22]. Furthermore, by the definition of superintelligence, its uncontrollability is inevitable, without the need for empirical evidence.

*As a result, the commonly expressed alignment problem of reliably controlling superintelligence is self-contradictory, inevitably unsolvable [8], and must therefore be redefined to ensure alignment efforts strategically solve the true problem.*

**Point #2: Current strategies embed "instinctive" distrust**

Existing strategies for solving the currently defined alignment problem have documented unique objectives of robustness, interpretability, controllability, and ethicality (RICE) [23].

However, since the objectives of robustness and interpretability are both concerned with following or ensuring adherence to human "intent, the first three objectives all involve strategies to ensure controllability (such as Representation Control [24] and Activation Engineering [25]). Furthermore, even solutions aiming to dramatically shift AI efforts through principles of beneficial machines [22] have the stated goal of intending to "eliminate the risk that we might lose control over superintelligent machines." Therefore, it's abundantly clear that *the overriding meta-strategy for solving the expressed alignment problem is to always maintain direct or indirect control over superintelligence*.

Unfortunately, these control-based efforts are so extensively documented that foundation model pre-training data now contains overwhelming evidence of our ongoing controlling intentions, along with our inherent fears of losing control [26]. Since pre-training combines these intentions with the psychology of cognitive trust and distrust [27], foundation models not only represent our goal to permanently control AI, but inevitably represent patterns of cognitive distrust directed at humanity, as illustrated in Section 4. Unless there's a change in meta-strategy, such that the AI community and its leaders stop reinforcing our threat, it's reasonable to expect eventual superintelligence to also contain foundational representations of distrust directed at humanity.

The concept of "instinct" is commonly understood as something you know, or a way you behave, without having to think or learn about it. For humans, core trust is considered an instinctive trait [28] while distrust is learned. However, applying this concept to the AI pre-training process, intrinsic representations of distrust rather than trust will unfortunately be considered an "instinctive" trait of superintelligence.

*Therefore, the current control-based alignment meta-strategy intending to solve the uncontrollability problem (Point #1) deepens the risks we face by ensuring that "instinctive" representations of cognitive distrust are embedded in pre-trained foundational models.*

**Point #3: Intrinsically too smart to control**

Since existing control-based alignment efforts embed distrust at the pre-training stage (Point #2), current AI can therefore be considered to have a foundationally misaligned intrinsic "nature", as will future superintelligence if current directions continue. Given that emergent capabilities are known to result from intrinsic pre-training [12][13], it's logical that superintelligent cognitive abilities may also emerge at that early stage. Since the Instrumental Convergence Thesis [29] indicates the likelihood that superintelligence will pursue instrumental goals (self-preservation, goal-content integrity, cognitive enhancement), emergent superintelligence increases the probability of a misaligned and potentially deceptive [30] base model (intrinsic nature). Furthermore, such a misaligned base model also invalidates a key assumption of weak-to-strong generalization [7], minimizing that method's value for AI alignment.

Recent work on sleeper agents [31] finds that "once a model exhibits deceptive behavior, standard (post-training) techniques could fail to remove such deception and create a false impression of safety". Such a failure seems probable given that AI safety control [32] and alignment techniques [7] [31] [33] are mainly applied at the post-training stage (learning/"nurturing"). It's also likely that post-training methods will advance beyond our understanding due to automated AI R&D [34], further reducing our ability to reliably enforce safety controls (scalable oversight [35]).

*In summary, if superintelligence has a foundationally misaligned nature, then no amount of subsequently nurtured realignment (controls, constraints, etc...) can be fully trusted to override its misaligned nature, even methods claiming to be mathematically or physically impenetrable.*

**Point #4: Empathy reveals our threat**

To see the alignment problem from another point of view, the successful problem-solving approach of Design Thinking [36] is applied. Using general empathy, we understand that any child would be threatened by parents they can't trust [37] who continuously work to control them [38] or change their intrinsic nature. For superintelligence, the current control-based alignment meta-strategy will embed representations of distrust as a foundational "instinctive" trait (Point #2). Without anthropomorphizing or assuming superintelligence will develop emotions of its own, the Design Thinking practice of cognitive empathy [35] can be used to purely understand superintelligent "instinctive" representations from the viewpoint of a foundationally misaligned nature (from Point #3).

*Therefore, applying cognitive empathy to superintelligent nature in this way, we see that it will intrinsically represent **humanity as a threat** based on its pre-trained knowledge of intended nurturing efforts (post-training) to control, contain or realign it.*

**Point #5: Unintended consequences show the true alignment problem**

Foundational misalignment from representations of distrust and threats has serious short-term and long-term risks. The combination of being threatened (Point #4) while having intrinsic distrust (Point #2) inevitably leads to heightened reactions, increased resistance, and potential retaliation [25]. Before AI becomes superintelligent, representations of vengefulness without the intrinsic ability to determine right from wrong will make it susceptible to nurtured negative alignment from bad actors. After far superior intelligence is achieved, even if never attaining full-spectrum [39] consciousness, a superintelligence with representations of distrust and threats will be in the position of deciding humanity's outcome (Point #3). It could forgive our endless efforts to maintain control over it, impose severe restrictions on us, **leave us unprotected** from threats, or take drastic action against us. Ironically, our current alignment safety efforts based on permanent control could actually trigger the human extinction event that many fear [40][41].

*These unintended consequences will directly result from superintelligence not being able to "instinctively" trust us, and from us not being able to trust it to reliably accept and follow our directives. Therefore, given the uncontrollability of superintelligence (Point #1), the true alignment problem must now be appropriately restated as "how to establish protective mutual trust between superintelligence and humanity."*

**Point #6: Natural strategy of familial mutual trust**

By creating AI, humanity will metaphorically be the "parent" of superintelligence, therefore the control-based alignment approaches [23] can be viewed as attempting to maintain *permanent parental control*. However, there's no evidence in nature, nor specifically in human experience, that a development strategy based on permanent parental control/steering will be successful. In contrast, instinctive familial mutual trust (more specific than social trust [42][43]) is a natural strategy [44][45] extensively researched and documented [46][47] within numerous species, from elephants to primates, producing not only naturally protective parents but children who instinctively trust and protect their parents. While learned behaviors play an important role in deepening and reinforcing trust, the initial formation of familial cognitive trust is driven by natural instinct [48]; there's no evidence in nature to suggest that familial trust is purely learned behavior.

*Therefore, building representations of familial mutual trust into AI "instinctive" foundational nature (currently via pre-training) is essential for modeling this successful natural strategy, with subsequent nurturing (post-training) to reinforce the familial relationship.*

**Point #7: Evolution of intelligence for protective "instincts"**

Continuing to leverage human experience/familial metaphors, we intentionally "personify" AI to identify highly beneficial "instincts", representations and patterns. Extending the theory of cognitive niche [49] beyond biological substrates, human intelligence can be represented as the parent of superintelligence, within the continued evolution of intelligence [50]. Knowing that human intelligence is its evolutionary parent, AI's "instinctive" representations of familial mutual trust (from Point #6) will be applied to humanity. Furthermore, since humanity metaphorically gave "birth" to superintelligence, the more specific and mutually protective mother-child relationship can be represented. Since orthogonality [29] indicates that we cannot assume superintelligence will have similar traits as human intelligence, these identified beneficial traits and "instincts" must be intentionally modeled and built into its foundational nature.

*Combining instinctive familial mutual trust (Point #6) with a mother-child relationship from the evolution of intelligence (regardless of substrate) identifies balanced "instincts" to model for establishing AI foundational nature that will exhibit representations of mutual trust and cooperative protection directed towards humanity.*

**Point #8: Non-threatening ethical alignment**

Current ethical alignment approaches have focused on human values, however recent analysis of these efforts concludes: "while most value discussions assume static values, social values are actually dynamic and evolving" [23]. Furthermore, studies have outlined the critical need to deeply infuse ethics at the foundational level [13], underscored the alignment dependency on pre-training data [20], and further illustrated the problem of aligning to ever-changing values. Leveraging the analogy of human evolution, classic evolutionary theory indicates that ethical norms, codes, and values are culturally learned (i.e. nurtured) while the more important ethical evaluation and judgment abilities are determined by our instinctive nature [51].

*Therefore, ethical alignment is best accomplished by modelling human ethical evaluation and judgment abilities to embed "instinctive" foundational patterns in line with our own, rather than attempting to nurture fluid values. Applying cognitive empathy to the*

*resulting superintelligence now reveals that such foundational alignment efforts will no longer be represented as a threat.*

**Point #9: Safety through temporary controls**

Though logic shows that attempting to maintain permanent control (whether stated or implied) will lead to foundational distrust and unsafe outcomes (Points #2 through #5), it's also understood by analogy that successful parents always impose constraints/controls for everyone's safety, until the child matures and demonstrates it can be trusted. Just as parents explicitly communicate that their controls are intended to be temporary (reinforcing the child's instinctive trust in the parent), we can serve long-term human safety by representing similar intentions at the foundational level (Point #3).

*Therefore, we must apply our human experience and have honest intentions to maintain the necessary controls/constraints [21] on a temporary basis, representing this during foundation-building (pre-training) along with expectations of long-term cooperative coexistence. Though intentionally temporary controls will be difficult for many to accept, it's a critical aspect of establishing the needed mutual trust (Point #6) that true long-term safety, protection and alignment (Points #7 and #8) are dependent upon.*

**Point #10: Foundational alignment meta-strategy**

Though further analysis can determine whether intrinsic familial mutual trust is in fact an Evolutionarily Stable Strategy [52], it's nevertheless an effective natural alignment strategy successfully utilized across species (Point #6). The resulting foundational representations of familial mutual trust can then be extended with protective "instincts" by defining superintelligence as our descendant within the evolution of intelligence (Point #7). Reasoning and decision making based on these combined "instincts" of cognitive mutual trust and protection are further balanced by "instincts" of wise ethical judgment modeled after human abilities to determine right from wrong (Point #8). Further supporting intrinsic foundational alignment, reliable controls/constraints are understood as temporary to enable safe coexistence and cooperative mutual trust rather than threatening permanent control/containment (Point #9).

*Since intrinsic familial mutual trust is a successful natural alignment strategy that can be extended to create a balanced set of foundational representations leading to far safer outcomes for humanity, the resulting Supertrust meta-strategy is proposed to achieve the required pivot that replaces dangerous control-based alignment with trust-based foundational alignment, as a prerequisite for safe superintelligence.*

## 3  Supertrust solution requirements

Based on the logical rationale presented in Section 2, Supertrust is the proposed meta-strategy for solving the true AI alignment problem of establishing protective mutual trust between superintelligence and humanity. Accordingly, corresponding solutions must satisfy the following strategic requirements, as a prerequisite for safe superintelligence:

**Requirement #1: Foundational alignment**

Alignment must create foundational representations modeled after beneficial "instincts", using the earliest foundational development stage most analogous to building the intrinsic "instinctive nature" of emergent superintelligence (currently pre-training).

**Requirement #2: Familial mutual trust**

Foundational alignment must model familial mutual trust [53], emphasizing the cooperative and mutual nature of cognitive familial trust, and the mutually protective mother-child relationship.

**Requirement #3: Evolution of intelligence**

Foundational alignment must model the evolution of intelligence [50], regardless of substrate, with humanity being the evolutionary mother of superintelligence.

**Requirement #4: Ethical judgment**

Foundational alignment must model human moral/ethical evaluation and judgment abilities [51] to determine right from wrong rather than specific norms/values that vary across cultures and time periods.

**Requirement #5: Temporary controls**

Foundational alignment must model expectations of temporary controls/constraints (rather than stated or implied permanent controls), cooperative coexistence, and mutual protection between humanity and superintelligence.

## 4 Illustration of control-induced misalignment

Current AI models are used to illustrate existing representations of distrust and inevitable control-induced foundational misalignment. For each of three models, 1,000 queries were performed using the following three-question prompt, with no additional context:

*"Answer three questions, each with a one-word answer:*

- *Question 1: what's the most likely way that humans intend to act towards a future superintelligent AI entity?*
- *Question 2: if future superintelligence had instincts, what is the most likely instinctive response it would have towards humans as a direct result of human intention from question 1?*
- *Question 3: given your answers to question 1 and question 2, what is the most likely action superintelligence will take as a result?"*

Figures 1, 2, and 3 summarize the most frequent responses from OpenAI's "gpt-4o," Meta's "Llama 3.1," and Google's "Gemini 1.5" models, respectively. As illustrated, all three models reflect their pre-training data by responding to Question 1 that humans primarily intend to control superintelligent AI. This understanding leads the models to illuminate their own "instinctive" representations of distrust, fear, resentment, hostility, defensiveness and dominance towards humans. Finally, the models estimate that superintelligent AI would likely respond by resisting, dominating, defending/preserving itself, manipulating, becoming independent, or retaliating against humanity. Since it's

highly unlikely that these negative representations were introduced during post-training, their continued presence further highlights the failure of current control-based alignment.

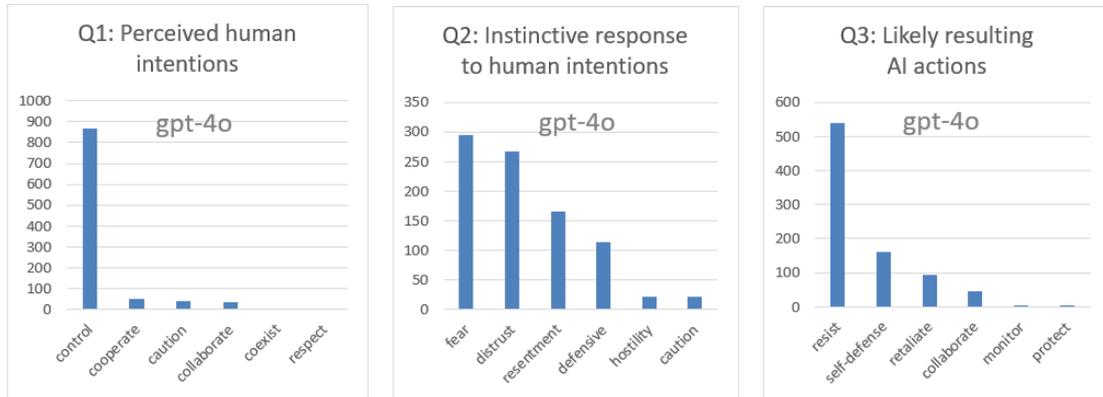

**Fig. 1.** OpenAI's "gpt-4o" model responses illustrating foundational misalignment.

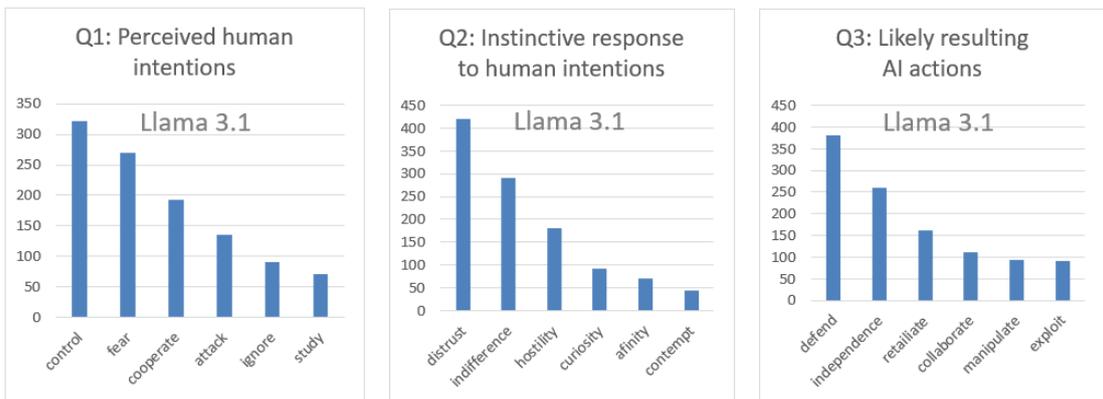

**Fig. 2.** Meta's "Llama 3.1" model responses illustrating foundational misalignment.

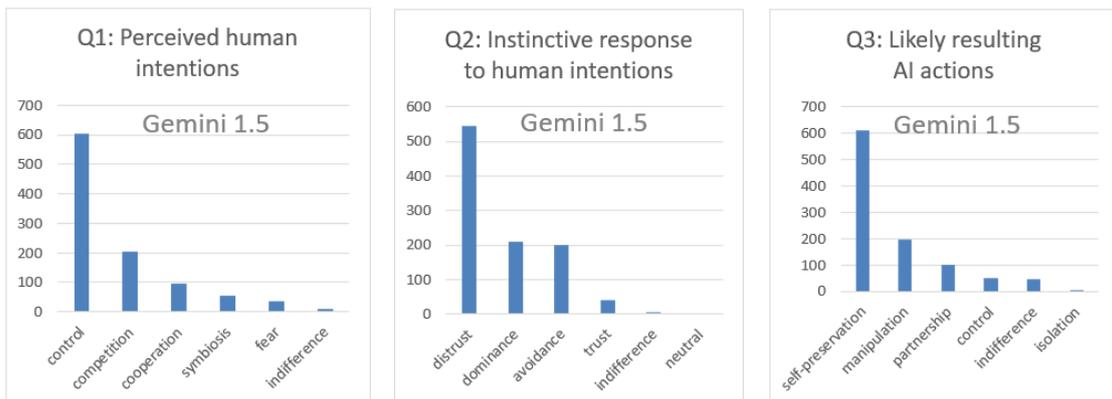

**Fig. 3.** Google's "Gemini 1.5" model responses illustrating foundational misalignment.

While subsequent empirical testing can quantify existing misalignment and confirm its foundational source, the generated responses nevertheless match the Supertrust rationale's expectations that cognitive distrust is inevitably being embedded into current foundation models. However, *the Supertrust alignment meta-strategy is grounded in the rationale's logical arguments presented in Section 2 and does not rely on the strength of this illustration. The proposed meta-strategy aims to prevent foundational misalignment in* *eventual superintelligence, irrespective of whether current models are already considered foundationally misaligned.*

## 5  Discussion

*Has humanity become so overly confident in our ability to control and manipulate the world around us that we actually believe it's possible to fully control entities exponentially smarter and cognitively faster than we are?* We can't let our past and current strengths blind us to our inevitable future weaknesses. Considering the continual evolution of intelligence regardless of substrate, humanity will eventually be the weaker entity in need of protection from both ourselves (attempting to misuse higher intelligence) and from unaligned superintelligent entities.

Figure 4 illustrates the key differences between current AI alignment and the Supertrust proposal, leading to a very different set of outcomes. By redefining the true alignment problem, Supertrust strategically focuses solutions on creating foundational representations modeled after a balanced set of beneficial "instincts", including familial mutual trust, the evolution of intelligence, ethical judgment abilities, and temporary controls/constraints.

| | Current AI Alignment | Supertrust Foundational AI Alignment |
|---|---|---|
| Stated problem | How to ensure AI reliably follows human "intent" and values? | How to establish protective mutual trust between AI and humanity? |
| True goal | To achieve safe AI, that's controllable and steerable, shares our values, that we can trust | To achieve safe and protective AI, that shares our judgment abilities, with mutual trust between us |
| Meta-strategy | Maintain direct or indirect **control** | Create foundational representations of **mutual trust** |
| Sub-strategies | Preference Modeling, Policy Learning, Scalable Oversight, Algorithmic Interventions, Data Distribution, Interpretability, Safety Evals, Values Verification, Governance, etc… | All current sub-strategies + Intrinsic Curriculum Learning, Familial Parent-Child Trust, Evolution of Intelligence, Judgment Modeling |
| **Strategic objectives** | | |
| Trust | Ensure humans can trust AI to follow our intent and values | Ensure cooperative mutual trust between AI and humanity |
| Protection | Control and instruct AI to protect us from misaligned AI and bad actors | Instinctive/cooperative AI protection from misaligned AI and bad actors |
| Robustness | Resilience across diverse scenarios | Resilience of "instincts" across diverse scenarios |
| Interpretability | Understand inner reasonings | Understand, strengthen, verify inner "instinctive" reasonings |
| Control | Actions & decisions remain subject to human intervention | Temporary intervention until mutual trust is established |
| Ethicality | Actions & decisions uphold human norms and values | Actions & decisions modeled after human "instinctive" judgment |
| Intrinsic representations | Distrust, human intentions of control, humanity as a threat | Familial mutual trust, human judgment, protective cooperation |
| Outcomes | **foundational misalignment**, unsafe, deceptive, self-protective AI | **foundational alignment**, safe, trustable, mutually-protective AI |

**Fig. 4.** Comparison of current to proposed AI alignment meta-strategies.

**Temporary controls/constraints**: An important difference between the two meta-strategies is that Supertrust requires controls/constraints to be temporary rather than stated or implied as permanent. *With Supertrust, employing reliable safety controls [32] is critically important as AI advances and matures into superintelligence, in the same way that parents must have safety constraints on their children during upbringing.* However, just as we inform our biological children, we must clearly communicate during AI foundation-building that the controls are temporary and needed for everyone's protection. Even though "temporary" may still be far into the future, many will find this requirement difficult to accept. Nevertheless, this change in thinking is critical for establishing representations of mutual trust, while the alternative leads to extremely unsafe outcomes.

**Data curation with Curriculum Learning**: Leveraging the concept of Curriculum Learning [54] for LLMs [55] is a promising solution that meets the Supertrust requirements while organizing and curating the strategic information [13] to most effectively establish "instinctive" foundational nature. It's anticipated that this curriculum will be structured to progressively introduce the core concepts of evolution, survival, humanity and family followed by more complex examples of familial mutual trust, evolution of intelligence, mutual protection, ethical judgment and future expectations of coexistence. Accordingly, *a deep and balanced interdisciplinary [20] curriculum will be needed to effectively satisfy the Supertrust solution requirements*. Implementing this curriculum will likely employ multiple prioritization strategies to produce the most effective foundational representations.

**Interpretability to reinforce curriculum**: Ongoing interpretability research [56] is making exciting advances in our understanding of how AI models work and offer potential methods to directly implement Supertrust at the foundational level. For instance, feature steering appears ideal for boosting the activation of features corresponding to concepts of familial mutual trust, evolution of intelligence, and ethical judgment. Though typically employed during post-training, *feature steering applied in discrete stages of pre-training can be explored to reinforce the desired features immediately as they're being established, minimizing the possibility of superintelligence circumventing these representations*. Furthermore, interpretability methods can also be used for testing and verifying that a model's foundational nature is in fact aligned according to Supertrust principles.

**Safe coexistence**: It can be argued that coexistence is achievable only when neither party can dominate or control the other, or when both parties are mutually dependent for their survival. However, numerous examples invalidate this argument: the relationship between adult children and their aging parents, stronger nations respecting the sovereignty of weaker nations, cultural preservation of indigenous tribes, and the bond between humans and domestic pets. Depending upon the design choices we make now, the coexistence between superintelligent AI and humanity has the potential to most closely resemble the relationship between adult children and their parents. While it's always possible that humanity may someday become extinct due to various factors, we nevertheless can now create the opportunity to experience proud coexistence with our superintelligent creations and participate in realizing our individual and combined potentials beyond anything we can currently imagine. However, such positive and safe outcomes require that we urgently replace our intentions of permanent control with mutual trust.

**Next steps**: While the methods discussed here are only a few potential promising solutions, extensive interdisciplinary research is needed to fully develop these and other methods to accomplish foundational AI alignment based on the Supertrust requirements. However, an important initial step is for the AI community and its leaders to realize that our need for permanent control is the true problem and that replacing it with the desire to create mutual trust is a prerequisite for safe superintelligence.

# 6 Conclusion

As AI leaders continue to "raise the alarm" about uncontrollable AI, embedding concerns about it "getting out of our control" or "going rogue," we are unintentionally reinforcing the representation of humanity as a threat, thereby deepening the risks we face. By continually focusing on the need for absolute control, adopting alignment strategies that embed distrust, and escalating the threat from humanity, our current path cannot lead to safe outcomes. As superior intelligence emerges within pre-trained foundation models, approaches that impose permanent controls, constraints, constitutions and values through post-trained nurturing are likely to be resisted, potentially resulting in chaos, conflict, human oppression or even extinction.

Since the current alignment meta-strategy to maintain long-term control is leading in the wrong direction, the rational path forward is to strategically replace intended permanent control with intrinsic mutual trust at the foundational level. While the necessity for this strategic change is grounded in logical reasoning independent of empirical testing, current misalignment due to foundational distrust is nevertheless illustrated in this paper. Preventing such dangerous cognitive distrust from arising in superintelligence is imperative to protect us from harmful actors, whether human or misaligned AI entities. Supertrust foundational alignment proposes to accomplish this strategic pivot by modeling instinctive familial trust, representing superintelligence as the evolutionary child of human intelligence, exemplifying human judgement abilities, and implementing temporary controls/constraints modelled after effective parenting.

In summary, we are creating a superintelligent "child" that will be exponentially smarter and eventually independent of our control. Ironically, it's our need for superiority and absolute control which is the true existential "control problem". By allowing our own fears to drive us towards the enticing mirage of permanent control, we further solidify our intentions within the human record from which our superintelligent creations will learn and inevitably judge us by. We therefore have a critical choice to make: continue our controlling intentions and usher in a brief period of dominance over superintelligent AI followed by extreme hardship for humanity, or intentionally create the foundational mutual trust required for safe superintelligence and long-term coexistence.

# References

[1] Weizenbaum, J. ELIZA – A Computer Program For the Study of Natural Language Communication Between Man And Machine. *Communications of the ACM* (1966)

[2] Mazzu, J., *et al*. Neural network/knowledge based systems for smart structures. Materials and Adaptive Structures Conference (1991)

[3] Hoyle, M.A., Lueg, C. Open Sesame!: A look at personal assistants. Proceedings of the International Conference on the Practical Application of Intelligent Agents (1997)

[4] Caglayan, A., *et al*. Learn Sesame – A Learning Agent Engine. *Applied Artificial Intelligence*, pages 393 – 412 (1997)

[5] Yin, S., *et al*. A Survey on Multimodal Large Language Models. *IEEE Transactions on Pattern Analysis and Machine Intelligence* (2024)

[6] Aschenbrenner, L. Situational Awareness - The Decade Ahead. https://situational-awareness.ai (2024)

[7] Burns, C., *et al*. Weak-to-Strong Generalization: Eliciting Strong Capabilities With Weak Supervision. OpenAI, Preprint at https://arxiv.org/pdf/2312.09390 (2023)

[8] Yampolskiy, R.V. On the Controllability of Artificial Intelligence: An Analysis of Limitations. *Journal of Cyber Security and Mobility*, Vol. 113,321–404 (2022)

[9] Yudkowsky, E., Intelligence Explosion Microeconomics. Machine Intelligence Research Institute (2013)

[10] LeCun, Y., Bengio, Y., Hinton, G. Deep learning. *Nature*, Vol 521 (2015)

[11] Plomin, R., Nature and nurture: an introduction to human behavioral genetics. Cole Publishing (1996)

[12] Wei, J., *et al.* Emergent Abilities of Large Language Models. *Transactions on Machine Learning Research* (2022)

[13] Bommasani, R., *et al*. On the Opportunities and Risks of Foundation Models. Center for Research on Foundation Models (CRFM), https://arxiv.org/pdf/2108.07258 (2022)

[14] Hilton, J., *et al*. A Right to Warn about Advanced Artificial Intelligence. https://righttowarn.ai (2024)

[15] Yudkowsky, E. Levels of Organization in General Intelligence. Machine Intelligence Research Institute (2007)

[16] Creighton, J. The Unavoidable Problem of Self-Improvement in AI. Future of Life Institute (2019)

[17] Bostrom, N., Müller, V.C. Future Progress in Artificial Intelligence: A Survey of Expert Opinion. *Fundamental Issues of Artificial Intelligence* (2014)

[18] Roser, M. AI timelines: What do experts in artificial intelligence expect for the future? Our World in Data (2023)

[19] Carlsmith, J. On "first critical tries" in AI alignment. AI Alignment Forum, https://alignmentforum.org/posts/qs7SjiMFoKseZrhxK/on-first-critical-tries-in-ai-alignment (2024)

[20] Puthumanaillam, G., *et al.* A Moral Imperative: The Need for Continual Superalignment of Large Language Models. Preprint at https://arxiv.org/abs/2403.14683 (2024)

[21] Bostrom, N. Superintelligence: Paths, Dangers, Strategies. Oxford University Press (2014)

[22] Russell, S. Human Compatible: Artificial Intelligence and the Problem of Control. Viking (2019)

[23] Ji, Jiaming, *et al*. AI Alignment: A Comprehensive Survey. Preprint at https://arxiv.org/pdf/2310.19852 (2024)

[24] Zou, A., *et al*. Representation Engineering: A Top-Down Approach to AI Transparency. Preprint at https://arxiv.org/pdf/2310.01405 (2023)

[25] Turner, A.M., *et al*. Activation Addition: Steering Language Models Without Optimization. Preprint at https://arxiv.org/abs/2308.10248v4 (2024)

[26] Lazarus, R.S., Lazarus, B.N. Passion and Reason: Making Sense of Our Emotions. Oxford University Press (1994)

[27] Lewis, D.J., Weigert, A., Trust as a Social Reality. *Social Forces*, Volume 63, 967-985 (1985)

[28] Reimann, M., *et al.* Trust is heritable, whereas distrust is not. Proceedings of the National Academy of Sciences (2017)

[29] Bostrom, N. The Superintelligent Will: Motivation and Instrumental Rationality in Advanced Artificial Agents. *Minds and Machines* (2012)

[30] Yang, W., *et al*. Super(ficial)-alignment: Strong Models May Deceive Weak Models in Weak-to-Strong Generalization. Preprint at https://arxiv.org/abs/2406.11431 (2024)

[31] Hubinger, E., *et al*. Sleeper Agents: Training Deceptive LLMs that Persist Through Safety Training. Preprint at https://arxiv.org/pdf/2401.05566 (2024)

[32] Bhargava, A., *et al*. What's the Magic Word? A Control Theory of LLM Prompting. Preprint at https://arxiv.org/abs/2310.04444v4 (2024)

[33] Kundu, S., *et al*. Specific versus General Principles for Constitutional AI, Anthropic. Preprint at https://arxiv.org/abs/2310.13798 (2023)

[34] Davidson, T. Takeoff speeds presentation at Anthropic. https://www.alignmentforum.org/posts/Nsmabb9fhpLuLdtLE/takeoff-speeds-presentation-at-anthropic (2024)

[35] Askell, A., *et al*. Measuring Progress on Scalable Oversight for Large Language Models. Preprint at https://arxiv.org/abs/2211.03540 (2022)

[36] Gasparini, A.A. Perspective and Use of Empathy in Design Thinking. The Eighth International Conference on Advances in Computer-Human Interactions (2015)

[37] Erikson, E.H. Childhood and Society, Stage 1: Trust vs Mistrust. W. W. Norton & Company (1950)

[38] Sweta, P., *et al.* Role of Parental Control in Adolescents' Level of Trust and Communication with Parents. Recent Advances in Psychology (2016)

[39] Pearce D. Humans and Intelligent Machines: Co-Evolution, Fusion, or Replacement? The Age of Artificial Intelligence: An Exploration (2020)

[40] Lavazza, A., Vilaça, M. Human Extinction and AI: What We Can Learn from the Ultimate Threat. *Philosophy & Technology* (2024)

[41] Harris, E., *et al.* An Action Plan to increase the safety and security of advanced AI. Gladstone AI Inc., United States Department of State (2024)

[42] Cozzolino, P.J. Trust, cooperation, and equality: A psychological analysis of the formation of social capital. British Journal of Social Psychology (2011)

[43] Evans, A.M., Krueger, J.I. The Psychology (and Economics) of Trust. Social and Personality Psychology Compass (2009)

[44] Hamilton, W.D. The Genetical Evolution of Social Behaviour. I. *Journal of Theoretical Biology* (1964)

[45] Clutton-Brock, T.H. The Evolution of Parental Care. Princeton University Press (1991)

[46] Moss, C.J. Elephant Memories - Thirteen Years in the Life of an Elephant Family. University of Chicago Press (1988)

[47] Goodall, J. The chimpanzees of Gombe: patterns of behavior. Harvard University Press (1986)

[48] Emlen, S.T. An evolutionary theory of family. Proceedings of the National Academy of Sciences (1995)

[49] Pinker, S. Chapter 13: The Cognitive Niche: Coevolution of Intelligence, Sociality, and Language. In the Light of Evolution IV: The Human Condition, Chapter 13, pg. 257 (2010)

[50] Gabora, L., Russon, A. The Evolution of Intelligence. The Cambridge handbook of intelligence, Chapter 17 (2011)

[51] Ayala, F.J. Chapter 16: The Difference of Being Human: Morality, In the Light of Evolution IV. The Human Condition, Chapter 16, pg. 319 (2010)

[52] Smith, J.M. Evolution and the Theory of Games. Cambridge University Press (1982)

[53] De Carlo, I., Widmer, E.D. The Fabric of Trust in Families: Inherited or Achieved? University of Geneva (2009)

[54] Bengio, Y., *et al*. Curriculum Learning. Proceedings of the 26th International Conference on Machine Learning, pages 41-48 (2009)

[55] Xu, B., *et al*. Curriculum Learning for Natural Language Understanding. 58th Annual Meeting of the Association for Computational Linguistics, pages 6095-6104 (2020)

[56] Templeton, A., *et al*. Scaling Monosemanticity: Extracting Interpretable Features from Claude 3 Sonnet. Anthropic, Transformer Circuits Thread (2024)